\documentclass{article} 
\usepackage{nips13submit_e,times}
\usepackage{hyperref}
\usepackage{url}
\usepackage{graphicx}
\usepackage{color}

\definecolor{dark_blue}{HTML}{3333cc}
\definecolor{dark_green}{HTML}{339933}

\title{Using General Adversarial Networks for Marketing: A Case Study of Airbnb}

\author{
Richard Diehl Martinez \\
Department of Computer Science\\
Stanford University\\
Stanford, CA 94305 \\
\texttt{rdm@stanford.edu} \\
\And
John Kaleialoha Kamalu \\
Department of Computer Science\\
Stanford University\\
Stanford, CA 94305 \\
\texttt{jkamalu@stanford.edu} \\
}

\nipsfinalcopy 

\begin{document}

\maketitle

\begin{abstract}
In this paper, we examine the use case of general adversarial networks (GANs) in the field of marketing. In particular, we analyze how GAN models can replicate text patterns from successful product listings on Airbnb, a peer-to-peer online market for short-term apartment rentals. To do so, we define the Diehl-Martinez-Kamalu (DMK) loss function as a new class of functions that forces the model’s generated output to include a set of user-defined keywords. This allows the general adversarial network to recommend a way of rewording the phrasing of a listing description to increase the likelihood that it is booked. Although we tailor our analysis to Airbnb data, we believe this framework establishes a more general model for how generative algorithms can be used to produce text samples for the purposes of marketing.  

\end{abstract}

\section{Introduction}

The development of online peer-to-peer markets in the 1990s, galvanized by the launch of sites like eBay, fundamentally shifted the way buyers and sellers could connect [4]. These new markets not only leveraged technology to allow for faster transaction speeds, but in the process also exposed a variety of unprecedented market-designs [4]. \\ \\ Today, many of the most well-known peer-to-peer markets like Uber and Instacart use a centralized system that matches workers with assigned tasks via a series of complex algorithms [4]. Still, a number of other websites like Airbnb and eBay rely on sellers and buyers to organically find one-another in a decentralized fashion. In the case of these decentralized systems, sellers are asked to price and market their products in order to attract potential buyers. Without a large marketing team at their disposal, however, sellers most often rely on their intuitions for how to present their articles or listings in the most appealing manner. Naturally, this leads to market inefficiencies, where willing sellers and buyers often fail to connect due to an inadequate presentation of the product or service offered. 

\section{Background}
Fortunately, we believe that the introduction of unsupervised generative language models presents a way in which to tackle this particular shortcoming of peer-to-peer markets. In 2014, Ian Goodfellow et. al proposed the general adversarial network (GAN) [5]. The group showcased how this generative model could learn to artificially replicate data patterns to an unprecedented realistic degree [5]. Since then, these models have shown tremendous potential in their ability to generate photo-realistic images and coherent text samples [5].\\ \\
The framework that GANs use for generating new data points employs an end-to-end neural network comprised of two models: a generator and a discriminator [5]. The generator is tasked with replicating the data that is fed into the model, without ever being directly exposed to the real samples. Instead, this model learns to reproduce the general patterns of the input via its interaction with the discriminator. \\ \\
The role of the discriminator, in turn, is to tell apart which data points are ‘real’ and which have been created by the generator. On each run through the model, the generator then adapts its constructed output so as to more effectively ‘trick’ the discriminator into not being able to distinguish the real from the generated data. The end-to-end nature of the model then forces both the generator and discriminator to learn in parallel [7]. 
While GAN models have shown great potential in their ability to generate realistic data samples, they are notoriously difficult to train. This difficulty arises from two-parts: 1) First, it is difficult to tune the hyper-parameters correctly for the adversarial model to continue learning throughout all of the training epochs [5]. Since both the discriminator and generator are updated via the same gradient, it is very common for the model to fall into a local minima before completing all of the defined training cycles. 2) GANs are computationally expensive to train, given that both models are updated on each cycle in parallel [5]. This compounds the difficulty of tuning the model’s parameters. \\ \\
Nonetheless, GANs have continued to show their value particularly in the domain of text-generation. Of particular interest for our purposes, Radford et al. propose synthesizing images from text descriptions [3]. The group demonstrates how GANs can produce images that correspond to a user-defined text description. It thus seems feasible that by using a similar model, we can produce text samples that are conditioned upon a set of user-specified keywords. \\ \\
We were similarly influenced by the work of Radford et. al, who argue for the importance of layer normalization and data-specific trained word embeddings for text generation [9] and sentiment analysis categorization. These findings lead us to question whether it is possible to employ recurrent neural networks with long short-term memory gates, as defined by Mikolov et al., to categorize product descriptions into categories based on the product's popularity [6]. 

\section{Data}

The data for the project was acquired from Airdna, a data processing service that collaborates with Airbnb to produce high-accuracy data summaries for listings in geographic regions of the United States. For the sake of simplicity, we focus our analysis on Airbnb listings from Manhattan, NY, during the time period of January 1, 2016, to January 1, 2017. The data provided to us contained information for roughly 40,000 Manhattan listings that were posted on Airbnb during this defined time period. For each listing, we were given information of the amenities of the listing (number of bathrooms, number of bedrooms …), the listing’s zip code, the host’s description of the listing, the price of the listing, and the occupancy rate of the listing. Airbnb defines a home's occupancy rate, as the percentage of time that a listing is occupied over the time period that the listing is available. This gives us a reasonable metric for defining popular versus less popular listings.

\section{Approach}

Prior to building our generative model, we sought to gain a better understanding of how less and more popular listing descriptions differed in their writing style. We defined a home’s popularity via its occupancy rate metric, which we describe in the \textbf{Data} section. Using this popularity heuristic, we first stratified our dataset into groupings of listings at similar price points (i.e. \$0-\$30, \$30-\$60, \dots ). Importantly, rather than using the home’s quoted price, we relied on the price per bedroom as a better metric for the cost of the listing. Having clustered our listings into these groupings, we then selected the top third of listings by occupancy rate, as part of the ‘high popularity’ group. Listings in the middle and lowest thirds by occupancy rate were labeled ‘medium popularity’ and ‘low popularity’ respectively. We then combined all of the listings with high/medium/low popularity together for our final data set.

\subsection{Recurrent Neural Network with Long Short-Term Memory Gates}

Using our cleaned data set, we now built a recurrent neural network (RNN) with long short-term memory gates (LSTM). Our RNN/LSTM is trained to predict, given a description, whether a home corresponds to a high/medium/low popularity listing. The architecture of the RNN/LSTM employs Tensorflow’s Dynamic RNN package. Each sentence input is first fed into an embedding layer, where the input’s text is converted to a GloVe vector. These GloVe vectors are learned via a global word-word co-occurrence matrix using our corpus of Airbnb listing descriptions [8]. At each time step, the GloVe vectors are then fed into an LSTM layer. For each layer, the model forward propagates the output of the LSTM layer to the next time-step’s LSTM layer via a rectified linear unit (RLU) activation function. Each layer also pipes the output of the LSTM through a cross-entropy operation, to predict, for each time-step, the category of the input sequence. We finally ensemble these predictions, to create the model’s complete output prediction.

\begin{figure}[h]
\begin{center}
\fbox{\includegraphics[scale=0.5]{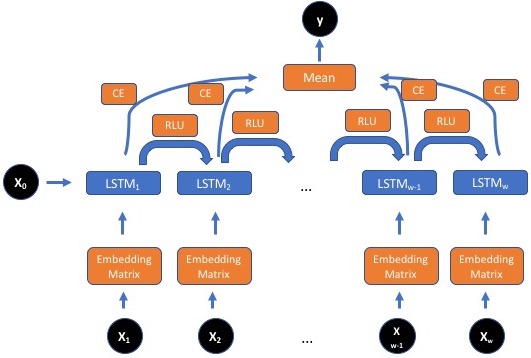}}
\end{center}
\caption{RNN/LSTM Framework}
\end{figure}

\subsection{Generative Adversarial Network}

\subsubsection{Model Framework} 
Having observed an unideal performance on this task (see Experiments below), we turned our attention to building a model that can replicate the writing style of high popularity listing descriptions. To solve this task, we designed a framework for a general adversarial network. This model employs the standard set up of a generator and a discriminator, but extends the framework with the adoption of the Diehl-Martinez-Kamalu loss. \\ \\
The generator is designed as a feed-forward neural network with three layers of depth. The input to the generator is simply a vector of random noise. This input is then fed directly to the first hidden layer via a linear transformation. Between the first and second layer we apply an exponential linear unit (ELU) as a non-linear activation function. Our reasoning for doing so is based on findings by Dash et al. that the experimental accuracy of ELUs over rectified linear units (RLU) tends to be somewhat higher for generative tasks [3]. Then, to scale the generator’s output to be in the range 0-1, we apply a sigmoid non-linearity between the second and the third layer of the model. \\ \\
The discriminator similarly used a feed-forward structure with three layers of depth. The input to the discriminator comes from two sources: real data fed directly into the discriminator and the data generated by the generator. This input is then piped into the first hidden layer. As before, an ELU transformation is then applied between the first and second layer, as well as between the second and third hidden layers. Finally, a sigmoid activation is used on the output of the last hidden layer. This sigmoid activation is important since the output of our discriminator is a binary boolean that indicates whether the discriminator believes the input to have been real data or data produced by the generator. This discriminator is thus trained to minimize the binary cross-entropy loss of its prediction (whether the data was real or fake) and the real ground-truth of each data point.

\begin{figure}[h]
\begin{center}
\fbox{\includegraphics[scale=0.5]{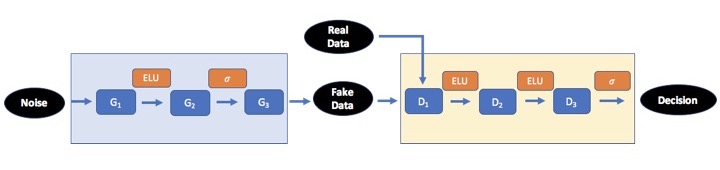}}
\end{center}
\caption{GAN Framework}
\end{figure}

The general framework defined above was inspired largely by the open-source code of Nag Dev, and was built using Pytorch [7]. One key extension to the basic GAN model, however, is the loss function that we apply to the generator, namely the Diehl-Martinez-Kamalu (DMK) Loss which we define below. 

\subsubsection{Diehl-Martinez-Kamalu Loss} 
The Diehl-Martinez-Kamalu Loss is a weighted combination of a binary cross entropy loss with a dot-product attention metric of each user-defined keyword with the model’s generated output. Formally, the binary cross entropy (BCE) loss for one example is defined as: 
$$ BCE(x,y) = y \cdot logx + (1-y) \cdot log(1-x), $$ where x is defined as the predicted label for each sample and y is the true label (i.e. real versus fake data). The DMK loss then calculates an additional term, which corresponds to the dot-product attention of each word in the generated output with each keyword specified by the user. 
To illustrate by example, say a user desires the generated output to contain the keywords, $\{subway, manhattan\}$. The model then converts each of these keywords to their corresponding glove vectors. Let us define the following notation $e(‘apple’)$ is the GloVe representation of the word apple, and let us suppose that $g$ is the vector of word embeddings generated by the generator. That is, $g_1$ is the first word embedding of the generator’s output. Let us also suppose $k$ is a vector of the keywords specified by the user. In our examples, $k$ is always in $R^{1}$ with $k_1$ one equaling of $‘subway’$ or $‘parking’$. The dot-product term of the DMK loss then calculates $\delta(g,k) = \sum_{i} \sum_{j} g_j \cdot e(k_i)$. Weighing this term by some hyper-parameter, $\gamma$, then gives us the entire definition of the DMK loss: \\ \\ \\ 
$$\delta(g,k) = \sum_{i} \sum_{j} g_j \cdot e(k_i)$$ \\ 
$$DMK(x,y,g,k) = BCE(x,y) - \gamma \delta(g,k)$$

\section{Experiments}

In seeking to answer the question of whether the occupancy rate of a listing could be extracted from the listing’s summary, we ran a number of experiments on our first model. Two parameterizations which we present here are (1) whether the word vectors used in the embedding layer are trained on our corpus or come pretrained from Wikipedia and Gigaword and (2) whether ensembling or the final hidden state in isolation are used to make a prediction for the sequence. Common to all experiments was our decision to use an Adam optimizer, 16 LSTM units, 50-dimensional GloVe vectors, and a 70-30 split in train and test data. 

\begin{figure}[h]
\begin{center}
\fbox{\includegraphics[scale=0.3]{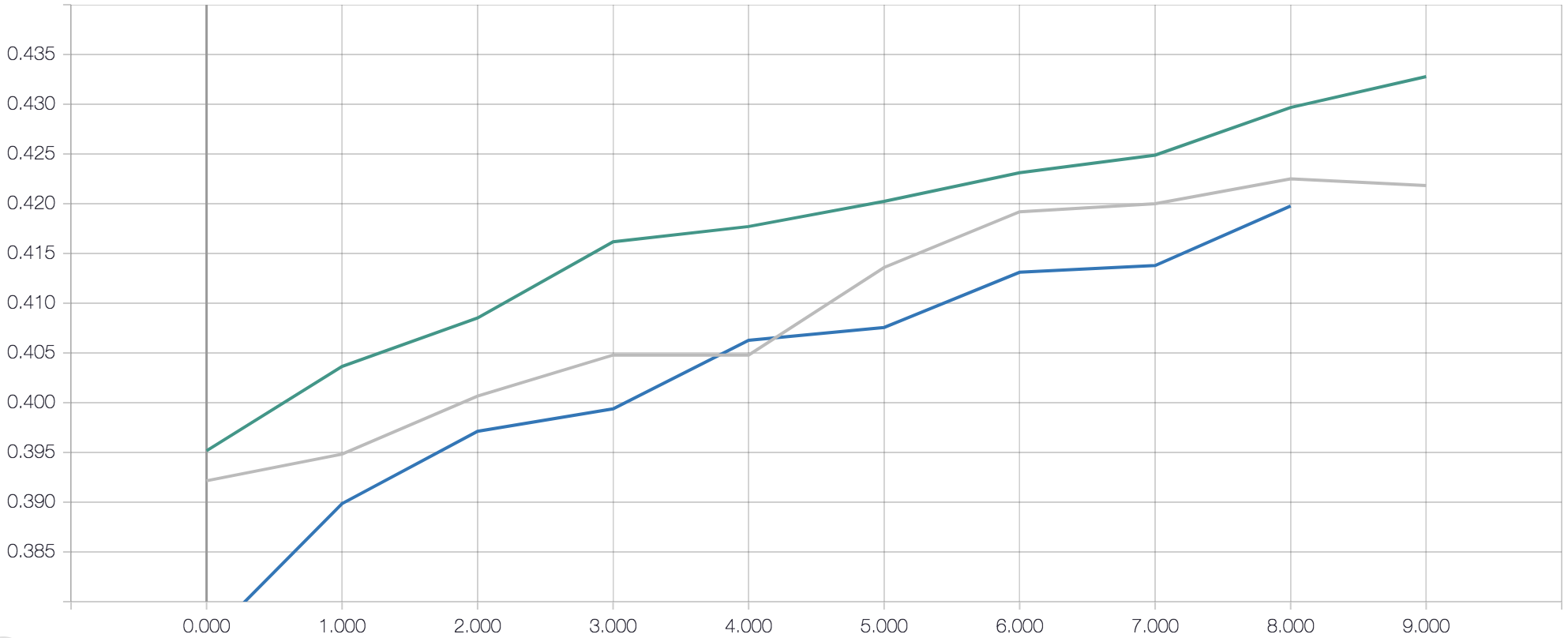}}
\end{center}
\caption{RNN/LSTM Accuracy over Number of Epochs}
\medskip
\begin{center}
\small
See Table 1 for a description of the models represented by the three colored lines. 
\end{center}
\end{figure}

\begin{figure}[h]
\begin{center}
\fbox{\includegraphics[scale=0.3]{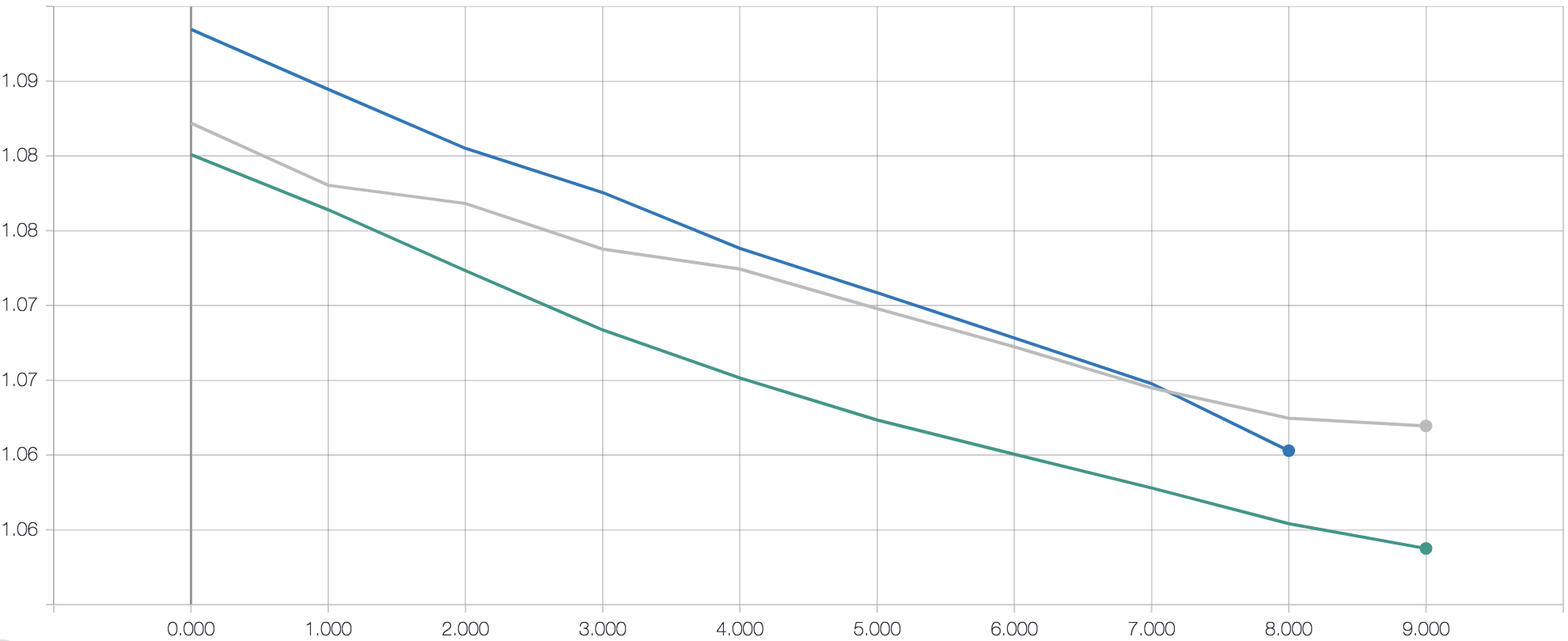}}
\end{center}
\caption{RNN/LSTM Loss over Number of Epochs}
\medskip
\begin{center}
\small
See Table 1 for a description of the models represented by the three colored lines. 
\end{center}
\end{figure}

Over ten epochs, the model parameterization which performs the best uses GloVe vectors trained on a corpus consisting of all listing descriptions and ensembling to make its class prediction. As a result, our findings are well in-line with those presented by Radford et. al who underscore the importance of training word embeddings on a data-specific corpus for best results on generative tasks [9]. 

\pagebreak

\begin{table}[h!]
\label{sample-table}
\begin{center}
\begin{tabular}{ll}
\multicolumn{1}{c}{\bf Model}  &\multicolumn{1}{c}{\bf Test Accuracy}
\\ \hline \\
\color{dark_green}{GloVe Vectors trained on Airbnb Data (Ensembling)}         & 0.403 \\
\color{gray}{GloVe Vectors trained on Airbnb Data (No Ensembling)}      & 0.397    \\
\color{dark_blue}{GloVe Vectors trained on Wikipedia Corpus (Ensembling)}    & 0.394
\end{tabular}
\caption{Results of RNN/LSTM}
\end{center}
\end{table} 

That said, these results, though they do show a marginal increase in dev accuracy and a decrease in CE loss, suggest that perhaps listing description is not too predictive of occupancy rate given our parameterizations. While the listing description is surely an influential metric in determining the quality of a listing, other factors such as location, amenities, and home type might play a larger role in the consumer's decision. We were hopeful that these factors would be represented in the price per bedroom of the listing -- our control variable -- but the relationship may not have been strong enough. \\ \\
However, should a strong relationship actually exist and there be instead a problem with our method, there are a few possibilities of what went wrong. We assumed that listings with similar occupancy rates would have similar listing descriptions regardless of price, which is not necessarily a strong assumption. This is coupled with an unexpected sparseness of clean data. With over 40,000 listings, we did not expect to see such poor attention to orthography in what are essentially public advertisements of the properties. In this way, our decision to use a window size of 5, a minimum occurrence count of 2, and a dimensionality of 50 when training our GloVe vectors was ad hoc. \\ \\
Seeking to create a model which could generate and discriminate a “high-occupancy listing description”, we wanted to evaluate the capabilities of a generative adversarial network trained on either the standard binary cross-entropy loss or the DMK loss proposed above. Common to both models was the decision to alternate between training the generator for 50 steps and the discriminator for 2000 steps. We leave further tuning of the models to future research as each occasionally falls into unideal local optima within 20 iterations. One potential culprit is the step imbalance between generator and discriminator -- should either learn at a much faster rate than the other, one component is liable to be ``defeated" and cease to learn the training data. \\ \\
Qualitatively the network trained on the DMK loss shows great promise. With respect to the two experiments presented here, we have shown that it is possible to introduce a measure of suggestion in the text produced by the generator. While this model is also subject to a rapid deadlock between generator and discriminator, it is interesting to see how the introduction of keywords is gradual and affects the proximal tokens included in the output. This behavior was made possible by paying close attention to the hyperparameter $\gamma$, the weight given to the dot product attention term of the DMK loss. After manual tuning, we settle on $\gamma=0.00045$ for this weight. Below, we illustrate model outputs using different values of Gamma. As is apparent, for a hyper-parameter value less than roughly $\gamma = 0.0004$, the model tends to ignore the importance of the keyword weights. Conversely, with a $\gamma$ value higher than $0.0005$, the model tends towards overweighting the representation of the keywords in the model output. 

\begin{table}[h!]
\label{sample-table}
\begin{center}
\begin{tabular}{ll}
\multicolumn{1}{c}{\bf $\gamma$} & \multicolumn{1}{c}{\bf Sample Output}
\\ \hline \\
0.0002   & our lovely unit now perfect on time before first same \\
& wonderful parking across orchard booking\\ \\
0.00045      & travelers apartment laundry street to block for the entire \\ & parking on laundry on accommodation street    \\ \\
0.0007   & and parking parking parking parking parking parking street \\
& parking parking convenient parking parking parking parking

\end{tabular}
\caption{GAN Model, Keywords = $[parking]$, Varying Gamma Parameter}
\end{center}
\end{table}

\section{Conclusion} 

We hope that this research paper establishes a first attempt at using generative machine learning models for the purposes of marketing on peer-to-peer platforms. As the use of these markets becomes more widespread, the use of self-branding and marketing tools will grow in importance. The development of tools like the DMK loss in combination with GANs demonstrates the enormous potential these frameworks can have in solving problems that inevitably arise on peer-to-peer platforms. Certainly, however, more works remains to be done in this field, and recent developments in unsupervised generative models already promise possible extensions to our analysis. \\ \\
For example, since the introduction of GANs, research groups have studied how variational autoencoders can be incorporated into these models, to increase the clarity and accuracy of the models’ outputs. Wang et al. in particular demonstrate how using a variational autoencoder in the generator model can increase the accuracy of generated text samples [11]. Most promisingly, the data used by the group was a large data set of Amazon customer reviews, which in many ways parallels the tone and semantic structure of Airbnb listing descriptions. More generally, Bowman et al. demonstrate how the use of variational autoencoders presents a more accurate model for text generation, as compared to standard recurrent neural network language models [1]. \\ \\
For future work, we may also consider experimenting with different forms of word embeddings, such as improved word vectors (IWV) which were suggested by Rezaeinia et al [10]. These IMV word embeddings are trained in a similar fashion to GloVe vectors, but also encode additional information about each word, like the word’s part of speech. For our model, we may consider encoding similar information in order for the generator to more easily learn common semantic patterns inherent to the marketing data being analyzed. 

\subsubsection*{References}
[1] Bowman, Samuel R., et al. "Generating sentences from a continuous space." \textit{arXiv:1511.06349} (2015). 

[2] Clevert, Djork-Arné, Thomas Unterthiner, and Sepp Hochreiter. "Fast and accurate deep network learning by exponential linear units (elus)."\textit{arXiv:1511.07289} (2015). 

[3] Dash, Ayushman, et al. "TAC-GAN-text conditioned auxiliary classifier generative adversarial network." \textit{arXiv:1703.06412} (2017). 

[4] Einav, Liran, Chiara Farronato, and Jonathan Levin. "Peer-to-peer markets." \textit{Annual Review of Economics 8} (2016): 615-635. 

[5] Goodfellow, Ian. "NIPS 2016 tutorial: Generative adversarial networks." \textit{arXiv:1701.00160} (2016). 

 [6] Mikolov, Tomas, Armand Joulin, Sumit Chopra, Michael Mathieu, and Marc Aurelio Ranzato. “Learning longer memory in recurrent neural networks.” \textit{arxiv:1412.7753}(2014).

[7] Nag, Dev. “Generative Adversarial Networks (GANs) in 50 Lines of Code (PyTorch).” \textit{Medium}, 10 Feb. 2017,  medium.com/@devnag/generative-adversarial-networks-gans-in-50-lines-of-code-pytorch-e81b79659e3f.

[8] Pennington, Jeffrey, Richard Socher, and Christopher Manning. "Glove: Global vectors for word representation." \textit{Proceedings of the 2014 conference on empirical methods in natural language processing (EMNLP)}. 2014. 

[9] Radford, Alec, Rafal Jozefowicz, and Ilya Sutskever. "Learning to generate reviews and discovering sentiment." \textit{arXiv:1704.01444} (2017).

[10] Rezaeinia, Seyed Mahdi, Ali Ghodsi, and Rouhollah Rahmani. "Improving the Accuracy of Pre-trained Word Embeddings for Sentiment Analysis." \textit{arXiv:1711.08609} (2017).

[11] Wang, Heng, Zengchang Qin, and Tao Wan. "Text Generation Based on Generative Adversarial Nets with Latent Variable." \textit{arXiv:1712.00170} (2017).

\end{document}